\documentclass{article}

\usepackage{arxiv}

\usepackage[utf8]{inputenc} % allow utf-8 input
\usepackage[T1]{fontenc}    % use 8-bit T1 fonts
\usepackage{hyperref}       % hyperlinks
\usepackage{url}            % simple URL typesetting
\usepackage{booktabs}       % professional-quality tables
\usepackage{amsfonts}       % blackboard math symbols
\usepackage{nicefrac}       % compact symbols for 1/2, etc.
\usepackage{microtype}      % microtypography
\usepackage{graphicx}
\usepackage{doi}

\usepackage{amsmath,amssymb}
\usepackage{algorithm}
\usepackage{algpseudocode}
\usepackage{svg}
\usepackage{xcolor}
\usepackage{xargs}

\title{Hierarchical Conditional Tabular GAN for multi-tabular synthetic data generation}

\author{\hspace{1mm}Wilhelm Ågren, Victorio Úbeda Sosa\\
    wilhelmagren98@gmail.com, victorio@kth.se\\
    Stockholm, Sweden
}

\renewcommand{\headeright}{}
\renewcommand{\undertitle}{A novel training and sampling algorithm for complex relational datasets}
\renewcommand{\shorttitle}{}

\begin{document}
\maketitle

\begin{abstract}
    The generation of synthetic data is a state-of-the-art approach to leverage when access to real data is limited or privacy regulations limit the usability of sensitive data. A fair amount of research has been conducted on synthetic data generation for single-tabular datasets, but only a limited amount of research has been conducted on multi-tabular datasets with complex table relationships. In this paper we propose the algorithm HCTGAN to synthesize multi-tabular data from complex multi-tabular datasets. We compare our results to the probabilistic model HMA1. Our findings show that our proposed algorithm can more efficiently sample large amounts of synthetic data for deep and complex multi-tabular datasets, whilst achieving adequate data quality and always guaranteeing referential integrity. We conclude that the HCTGAN algorithm is suitable for generating large amounts of synthetic data efficiently for deep multi-tabular datasets with complex relationships. We additionally suggest that the HMA1 model should be used on smaller datasets when emphasis is on data quality.
\end{abstract}


\section{Introduction}
\label{sec:Introduction}

Data is ubiquitous in today's modern society, and being able to opt-out of sharing your data with services is made more and more difficult each day. Nevertheless, we have seen great initiatives and regulations in the last decade, like the GDPR \cite{GDPR2016a} in the EU and PIPA \cite{pipaKorea} in South Korea, whose goals are to enforce standards for individuals data privacy. But providing greater privacy standards for individuals does not come cheap for parties that aim to leverage the data, as new processes for using the data has to be established, and a lot of the data is now rendered unusable or inaccessible for development and analytical initiatives. Synthetic data is today being actively researched within the field of machine learning, and is an approach to leverage in businesses where usability of real data is limited. Furthermore, it has showed to be an effective method to use for speeding up internal data processes whilst minimizing privacy risks \cite{mikael2021syn, wu2024usingsyntheticdatamitigate}.

Tabular data is a de-facto standard used throughout the industry to store data in a column and row format. Typically, each row represents an object and each column represents an attribute of the object. The rows are usually identified by a unique identifier, ID, which is stored as a column in the table and is referred to as a primary key of the table. If data is stored in multiple tables which are connected, this is referred to as relational data, or multi-tabular data. To represent a relationship between two tables, the child table has to contain a reference to some object in the parent table. This is accomplished by introducing a new column in the child table which stores so called foreign keys. The foreign keys reference an object in the parent table, more specifically, they reference the parents primary key. Parent tables can have multiple children, and vice versa for children, which means that there could be highly complex relationships in a multi-tabular database.

Machine learning research on synthetic data has thus far showed promising results, but the large majority of that research only deals with non-relational data \cite{xu2018synthesizing, xu2019modeling, Park_2018}. Hence, there is a lack of established research on multi-tabular algorithms, and the little research that does exist  mainly proposes to use probabilistic models. However, these models do not scale well for largely connected multi-tabular datasets, neither in the training nor in the sampling process. Furthermore, there is no guarantee that the generated synthetic data has referential integrity between tables \cite{canale2022generative}. Apart from probabilistic models, plenty of research has been conducted on Generative Adversarial Networks (GANs) for synthetic data generation, but mainly on single table datasets \cite{gueye2022row}. The proposed GAN models are an attractive choice whenever sensitive data is to be synthesized, because the generator never operates on any real data, and these models can be extended to incorporate various levels of differential privacy \cite{decristofaro2020overviewprivacymachinelearning, xie2018differentially, dwork2006dp, wood_differential_2018}.

\section{Related work}
\label{sec:Related_work}

The GAN framework is comprised of two neural networks that are trained adversarially in a zero-sum game. The generator $\mathcal{G}$ is trained to decode latent space noise to resemble the real target data, and a discriminator $\mathcal{D}$ is trained to distinguish between the generated data and the real data \cite{goodfellow2014generative}. The framework has been researched extensively in a wide range of areas, e.g., speech generation with HiFi-GAN \cite{bińkowski2019high, kong2020hifigan}, image generation with DCGAN \cite{radford2016unsupervised, huang2018introduction}, image-to-image translation with SPA-GAN and Pix2Pix \cite{emami2020spagan, isola2018imagetoimage}, and has seen a number of extensions in the form of StackGAN and InfoGAN to name a few \cite{perarnau2016invertible, reed2016generative, odena2016semisupervised, larsen2016autoencoding, zhang2017stackgan, salimans2016improved}. The GAN framework has been employed for tabular data generation tasks, where an initial model TGAN showed promising results and has subsequently been iteratively improved \cite{xu2018synthesizing, kuo2020generative}.

The CTGAN model improves the TGAN model by introducing three novel concepts: mode-specific normalization, a conditional generator, and training by sampling. These concepts together drastically improves both model training and the quality of the synthesized tabular data, by dealing with the problems of class imbalance and complicated column distributions \cite{xu2019modeling}. The CTAB-GAN model improves CTGAN by including the so-called classification and information loss. These modifications are said to improve semantic column integrity and stability in training \cite{pmlr-v157-zhao21a}. These tabular models are based on the WGAN modifications and incorporates the Wasserstein metric in the loss function and modifies the discriminator to not discriminate instances. Instead, it is often referred to as a critic $\mathcal{C}$ because it outputs a number representing the credibility of the generated sample. The WGAN extension improves gradient behavior and simplifies the optimization process of the generator \cite{arjovsky2017wasserstein}. Techniques from the PacGAN framework were also utilized in the CTGAN model in order to prevent mode collapse, thus, improving the diversity of the generated samples \cite{NEURIPS2018_288cc0ff}.

The RC-TGAN model was proposed to deal with complex multi-tabular datasets by conditioning the input of the child table generation process on previously generated parent rows, specifically, the features of the parents. This is done in order to transfer relationship information from the parent to the child table.  However, conditioning the child generator on row features from the parent tables leads to the generators being exposed to real data, and thus, potentially vulnerable to leakage of sensitive data \cite{gueye2022row, ganev2023syntheticdatametregulation}. Utilizing transformers for sequence-to-sequence modeling of relational tabular data was recently proposed and shows promising results as a substitute to real data for machine learning tasks. The authors utilize the parent tables sequence-to-sequence network as encoder for training its child tables, speeding up training times. Furthermore, utilizing the trained parent encoder allows information learned about the parent table to transfer to the decoder of the child table. However, exposing the encoder and decoder networks to the real data samples again leaves the model vulnerable \cite{solatorio2023realtabformer, gao2024trialsynthgenerationsyntheticsequential, wu_methodology_2016}.

\section{HCTGAN}
\label{sec:HCTGAN}

The Hierarchical Conditional Tabular GAN (HCTGAN) is our extended version of the CTGAN model and introduces information transfer between the parent- and child table generator networks to promote relational learning. This work introduces two novel algorithms, one for training and one for sampling. Both algorithms are suitable for synthesizing large, and arbitrarily complex, multi-tabular datasets. The sampling algorithm retains the original database structure and ensures referential integrity in the synthesized data between relating tables. We will first present some technical background on the CTGAN model and then introduce the hierarchical elements which make up the HCTGAN model.

\subsection{CTGAN}
\label{sec:HCTGAN_CTGAN}
The generator and critic networks in the CTGAN model aim to respectively minimize and maximize the Wasserstein GAN loss $\mathcal{L_W}$ with gradient penalty and packing, which is formally described as 

\begin{equation}
    \underset{\mathcal{G}}{\text{min}}\ \underset{\mathcal{C}}{\text{max}}\ \mathcal{L}_\mathcal{W} = 
    \underbrace{\underset{\Tilde{x}\sim\mathbb{P}_\mathcal{G}}{\mathbb{E}}[\mathcal{C}(\Tilde{x})] - \underset{x\sim\mathbb{P}_r}{\mathbb{E}}[\mathcal{C}(x)]}_{\text{Critic loss}} +
    \underbrace{\lambda\underset{\Hat{x}\sim\mathbb{P}_{\Hat{x}}}{\mathbb{E}}[(\| \nabla_{\hat{x}}\ \mathcal{C}(\hat{x}) \|_2 - 1)^2]}_{\text{Gradient penalty}}
\end{equation}

where $\Tilde{x}\sim\mathbb{P}_{\mathcal{G}}$ is the generator distribution, $x\sim\mathbb{P}_r$ is the real data distribution, and $\hat{x}\sim\mathbb{P}_{\hat{x}}$ is the uniform distribution of samples between pairs of points from $\mathbb{P}_{\mathcal{G}}$ and $\mathbb{P}_r$ \cite{gulrajani2017improved}.

\textbf{Mode-specific normalization} is a technique used to model columns with complicated distributions \cite{deecke2018mode}. The number of modes $m_i$ for each continuous column $C_i$ is estimated with a variational Gaussian mixture model. The estimated number of modes $m_i$ is used to fit a Gaussian mixture for the column $C_i$ such that the learned mixture becomes

\begin{equation}
    \mathbb{P}_{C_i} = \sum_k^m\mathcal{N}(c_{i,j};\eta_k,\phi_k)\mu_k = \sum_k^m \rho_k
\end{equation}

where $c_{i,j}$ is value $j$ in $C_i$, $\eta_k$ is mode $k$ of the mixture model, $\rho_k$ is the probability density for the mode, $\mu_k$ and $\phi_k$ represents the mode weight and standard deviation respectively. One mode is sampled from the probability densities $\rho_k$ of the mixture model and is then used to normalize the value $c_{i, j}$ to produce the scalar

\begin{equation}
    \alpha_{i,j} = \frac{c_{i,j} - \eta_k}{4\phi_k}.
\end{equation}

The scalar is used together with the one-hot vector $\boldsymbol{\beta}_{i,j}$, which indicates the sampled mode, to represent the row data according to

\begin{equation}
    \mathbf{r}_j = \alpha_{1,j} \oplus \boldsymbol{\beta}_{1,j} \oplus \cdots \oplus \alpha_{N_C,j} \oplus \boldsymbol{\beta}_{N_C,j} \oplus \mathbf{d}_{1,j}\oplus\cdots\oplus\mathbf{d}_{N_d,j}   
\end{equation}

where categorical and discrete values are represented as the one-hot vector $\mathbf{d}_{i,j}$ and $\oplus$ denotes concatenation.

\textbf{Conditional generator} introduces three primary concepts: the conditional vector, the generator loss, and training-by-sampling. The conditional vector $\mathbf{c}_j$ is a masked concatenation of all categorical and discrete one-hot encoded vectors $\mathbf{d}_{i, j}$ according to

\begin{equation}
    \mathbf{c}_j = \mathbf{m}_j (\mathbf{d}_{1,j} \oplus \cdots \oplus \mathbf{d}_{N_d, j})
\end{equation}

where the mask vector $\mathbf{m}_j$ marks the index of column $i$ with value $k$ with a $1$ and $0$ for all other values, when subject to the condition $(D_{i^*} = k^*)$. For example, given the two discrete columns $D_1 = \{-2, 3, 4\}$ and $D_2 = \{\text{cat}, \text{dog}\}$ subject to the condition $(D_2 = \text{cat})$ the resulting conditional vector would become $c=[0, 0, 0, 1, 0]$. The sampling of the conditional vector is drawn from a log-frequency distribution of each category. Conditioning the input for $\mathcal{G}$ evenly on categorical and discrete values is necessary in order to learn the real conditional distribution as

\begin{equation}
\label{eq:CTGAN_Conditional-probability}
    \mathbb{P}(\text{row}) = \sum_{k\in D_{i^*}}\mathbb{P}_\mathcal{G}(\text{row}\ |\ D_{i^*} = k)\mathbb{P}(D_{i^*} = k).
\end{equation}

The generator loss is introduced to allow $\mathcal{G}$ to produce any set of discrete vectors $\mathbf{d}_{i^*}$ during training . Meaning, the condition $(D_{i^*} = k^*)$ can be violated such that $k \neq k^*$. The cross-entropy between $\mathbf{m}_j$ and the generated $\mathbf{d}_{i^*}$ is introduced to penalize the loss, driving $\mathcal{G}$ to make a direct copy of $\mathbf{c}_j$ into $\mathbf{d}_{i^*}$ \cite{xu2019modeling}.

\subsection{Hierarchical modeling}
\label{sec:HCTGAN_Hierarchical-modeling}

Let $\mathcal{P}(\mathcal{T}_i)$ denote the set of parents for table $\mathcal{T}_i$. Given a table $\mathcal{T}_i$ with no cyclic relations, and at least one relation, the HCTGAN algorithm generates synthetic data row-wise according to the following conditional probability distribution

\begin{equation}
\label{eq:HCTGAN_Conditional-probability}
    \mathbf{\hat{r}}_i \sim \mathbb{P}_{\mathcal{G}_i}(\text{row}\ |\ D_{i^*} = k^*,\ \mathbf{z}_i) 
\end{equation}

where $\mathbf{z}_i$ is a concatenation of column-wise Gaussian noise for each column in the parent tables, such that

\begin{equation}
\label{eq:hctgan_zi}
    \mathbf{z}_i \sim  \mathcal{N}(\boldsymbol{\mu}_{i-1}, \boldsymbol{\sigma}_{i-1}) =
    \begin{cases} 
     \underbrace{\mathcal{N}(\mu_1, \sigma_1) \oplus \cdots \oplus \mathcal{N}(\mu_n, \sigma_n)}_{n\ \text{parent columns}},\ \ \text{if }\ |\mathcal{P}(\mathcal{T}_i)| \geq 1 \\
     \mathcal{N}(0, 1),\ \ \ \ \ \ \ \ \ \ \ \ \ \ \ \ \ \ \ \ \ \ \ \ \ \ \ \ \ \ \ \ \ \ \ \ \ \ \ \text{otherwise}.
    \end{cases}
\end{equation}

Because $\mathcal{G}_i$ needs to condition on the categorical and discrete values evenly to learn the real data distribution, we adhere to the generator loss and training-by-sampling process proposed by CTGAN. However, in HCTGAN we also condition on the noise vector $\mathbf{z}_i$ as a means to introduce information from parent tables. Since the conditional vector $\mathbf{c}$ and $\mathbf{z}_i$ are independent, the joint conditional distribution in equation \ref{eq:HCTGAN_Conditional-probability} can be marginalized independently. This leads to two tractable probability distributions and when $\mathcal{G}_i$ conditions evenly on $(D_{i^*} = k)$ the real conditional distribution can be learned as in equation \ref{eq:CTGAN_Conditional-probability}.

\begin{figure}[h]
\label{fig:HCTGAN}
  \centering
  \includegraphics[width=12cm]{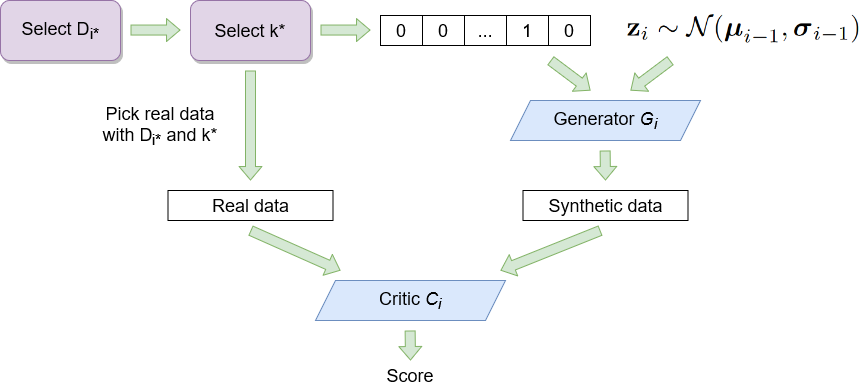}
  \caption{Schematic training overview of the HCTGAN model for one table $\mathcal{T}_i$. The generator $\mathcal{G}_i$ is conditioned on both the conditional vector $\mathbf{c}$ and noise $\mathbf{z}_i$ drawn from Gaussian distributions modeled after the parent table columns.}
\end{figure} 

The generators $\mathcal{G}_i$ follow a similar residual decoder network architecture as proposed in the CTGAN paper, with batch normalization and \texttt{ReLU} being applied before each residual connection, activation function \texttt{tanh} for numerical columns and \texttt{gumbel softmax} for categorical and discrete columns \cite{jang2017categorical}. The critics $\mathcal{C}_i$ also follow the same network architecture as proposed in the CTGAN paper, with dropout and \texttt{LeakyReLU} being applied after each affine layer \cite{xu2019modeling}. A schematic overview of the HCTGAN training process for a table $\mathcal{T}_i$ can be seen in figure \ref{fig:HCTGAN}, and pseudo code for the training and sampling algorithms can be found in appendix \ref{appendix:HCTGAN-training-algorithm} and \ref{appendix:HCTGAN-sampling-algorithm} respectively.

\section{Results}
\label{sec:Results}
The following sections describes the dataset and metrics used for evaluating our proposed HCTGAN algorithms, followed by the experimental results and a discussion.

\subsection{Datasets}
\label{sec:Results_Datasets}
Given that the HMA1 model scales poorly for datasets with many relational tables, the amount of datasets available for comparison is limited. Further research on larger datasets when solely using the HCTGAN model would be interesting to investigate in future work. We therefore used three datasets in the evaluation process which are manageable by the HMA1 algorithm and are publicly available through the SDV Python library \cite{sdvdatasets}:

\textbf{University\_v1} is a two-level dataset with 5 tables in total, containing three parent- and two child tables. The child tables both have two parents each, and one parent table has two children. They contain both one-to-one and one-to-many relationship types.

\textbf{Hepatitis\_std\_v1} is a two-level dataset with 7 tables in total, containing four parent- and three child tables. The child tables all have two parents each, one parent table has three children, one has two children, and the other two parents only have one child table. The tables contain both one-to-one and one-to-many relationship types.

\textbf{Pyrimidine\_v1} is a two-level dataset with 2 tables, one parent- and one child table. The relationship type is one-to-many.

\subsection{Metrics}
\label{sec:Results_Metrics}
Several established metrics from the SDMetrics Python library by SDV were used to evaluate the performance of our proposed algorithms \cite{sdmetrics}. These are defined as follows:

\textbf{KSComplement (CS)} computes the similarity between real $x$ and synthetic numerical data $\hat{x}$ by using the Two-sample Kolmogorv-Smirnov statistic $D$. Given the two estimated CDFs $F_n$ and $F_m$, where $n$ and $m$ are the CDFs respective sample sizes, the KS statistic is calculated accordingly

\begin{equation}
\label{eq:KS-stat}
    D_{n, m} = \underset{x, \hat{x}}{\text{sup}}\ | F_n(x) - F_m(\hat{x}) |
\end{equation}

where $\text{sup}_{x, \hat{x}}$ is the supremum of the set of absolute distances between the two CDFs. The KSComplement metric is normalized to the range $[0, 1]$, and the KS statistic $D$ is inverted such that a higher score means higher quality data.

\textbf{TVComplement (CS)} computes the similarity between real $x$ and synthetic categorical data $\hat{x}$ by utilizing the Total Variation Distance (TVD). The first step is to compute the frequency of all categories, and then one can calculate the TVD as follows

\begin{equation}
\label{eq:TVD}
    \delta(\omega, \hat{\omega}) = \frac{1}{2} \sum_{k\in D_i} | \omega_k - \hat{\omega}_k |
\end{equation}

where $k$ represents all possible categories in column $D_i$, $\omega$ and $\hat{\omega}$ represents the frequency of those categories for the real and synthetic data respectively. The TVD score is normalized to the range $[0, 1]$ and inverted such that a higher score means higher quality data.

\textbf{CorrelationSimilarity (CPT)} measures the correlation between a pair of numerical columns $A, B$ by computing either the Pearson- or Spearman Coefficient. The metric is calculated accordingly

\begin{equation}
\label{eq:CorrelationSimilarity}
    \text{CorrelationSimilarity}(x, \hat{x}) = 1 - \frac{| f(x_{A, B}) - f(\hat{x}_{A, B}) |}{2}
\end{equation}

where $f(x) \in [-1, 1]$ is the Pearson- or Spearman Coefficient. The metric is normalized to $[0, 1]$ where a higher score indicates that the pairwise correlations are similar.

\textbf{ContingencySimilarity (CPT)} measures the correlation between a pair of categorical columns $A, B$ by computing a normalized contingency table for the real $x$ and synthetic data $\hat{x}$. The table represents the frequencies of categorical combinations in $A$ and $B$. The total metric is calculated in the following way

\begin{equation}
\label{eq:ContingencySimilarity}
    \text{ContingencySimilarity}(x, \hat{x}) = 1 - \frac{1}{2}\sum_{a\in A}\sum_{b\in B} | \omega_{a, b} - \hat{\omega}_{a, b} |
\end{equation}

where $a$ and $b$ represents all possible categories in columns $A$ and $B$, $\omega$ and $\hat{\omega}$ represents the frequency of those categories for the real and synthetic data respectively. The metric is normalized to $[0, 1]$ where a higher score means that the contingency table is similar between the real and synthetic data.

\textbf{CardinalityShapeSimilarity (PCR)} measures how similar the cardinality of primary- and foreign keys are between real $x$ and synthetic data $\hat{x}$. Practically it computes the number of child foreign keys that each parent primary key is connected to, denoted by $n(x)$. Doing this for all primary- and foreign key relations yields a numerical distribution $F_i(n(x))$. The KSComplement is then used to measure how similar the real and synthetic data cardinality distributions are as follows

\begin{equation}
\label{eq:PCR}
    \text{PCR}(x, \hat{x}) = 1 - \underset{x, \hat{x}}{\text{sup}}\ | F_n(n(x)) - F_m(n(\hat{x})) |.
\end{equation}

We denote this metric as Parent Child Relationship (PCR) in the results. The metric is bounded by the range $[0, 1]$ where a high score indicates that the synthetic data relation cardinalities is similar to the real data.

\textbf{RangeCoverage (RC)} measure how well numerical synthetic data $\hat{x}$ covers the entire range of real values $x$, column-wise. The score is calculated with the following formula

\begin{equation}
\label{eq:RangeCoverage}
    \text{RangeCoverage}(x, \hat{x}) = 1 - \biggl[ \text{max} \biggl( \frac{\text{min}(\hat{x}) - \text{min}(x)}{\text{max}(x) - \text{min}(x)},\ 0 \biggr) + \text{max} \biggl( \frac{\text{max}(x) - \text{max}(\hat{x})}{\text{max}(x) - \text{min}(x)} ,\ 0 \biggr) \biggr].
\end{equation}

The metric can take on values below $0$ if the synthetic data has terrible range coverage. Nonetheless, the resulting score is thresholded to always be positive and in the range $[0, 1]$, where a higher score indicates that the synthetic data covers the entire range of numerical values.

\textbf{CategoryCoverage (RC)} measures how well categories are present in the categorical synthetic data $\hat{x}$ compared to the real data $x$, column-wise. The metric is calculated by computing the number of unique categories $\hat{\omega}$ in the synthetic column, and dividing by the number of unique categories $\omega$ in the real column. The score is defined on the range $[0, 1]$ where a high score indicating that the synthetic data covers the real data categories well.

\textbf{NewRowSynthesis (NRS)} measures the total amount of synthetic data rows which are novel by looking for matching rows. The metric works differently depending on the type of the data. For categorical data the values must exactly match, whereas for numerical data the values are scaled and considered a match whenever the real data is within some \% of the synthetic data (default is 1\%). The score is defined on the range $[0, 1]$ where a high score indicates that the synthetic data is novel.

\textbf{BoundaryAdherence (BA)} measures how well the numerical synthetic data $\hat{x}$ complies with the min- and max values of the real data $x$. The min- and max values of the real data are found, and the frequency of synthetic data $\hat{\omega}$ which lie within the valid range constitutes the metric. The metric takes values in the range $[0, 1]$ where a high score indicates that the synthetic data follows the min- and max values of the real data.

\subsection{Experimental results}
\label{sec:Results_Experimental-results}
The HMA1 model was fit
 according to its specified procedure, and the HCTGAN models were trained for 50 epochs using the Adam optimizer for all GANs with weight decay $\lambda = 1\text{e}-6$, learning rate $\eta=2\text{e}-4$, and decay rates $(\beta_1, \beta_2) = (0.5, 0.9)$ \cite{adamOpt}. The models were set to synthesize approximately the same amount of rows for each dataset respectively. All experiments were run for $n=3$ times using different seeds, and the presented metrics are aggregated mean and standard deviation of each experiment.

\begin{table}[H]
\centering
    \begin{tabular}{c | c | c}
        Metrics & HMA1 & HCTGAN \\
        \hline
        CS & \textbf{0.807} $\boldsymbol{\pm}$ \textbf{0.0} & 0.716 $\pm$ 0.0235 \\
        CPT & \textbf{0.677} $\boldsymbol{\pm}$ \textbf{0.0} & 0.501 $\pm$ 0.0099\\
        PCR & \textbf{0.715} $\boldsymbol{\pm}$ \textbf{0.0} & 0.430 $\pm$ 0.0023 \\
        RC & 0.944 $\pm$ 0.0 & \textbf{1.0} $\boldsymbol{\pm}$ \textbf{0.0} \\
        NRS & 0.903 $\pm$ 0.0 & \textbf{0.964} $\boldsymbol{\pm}$ \textbf{0.0068} \\
        BA & \textbf{1.0} $\boldsymbol{\pm}$ \textbf{0.0} & 0.737 $\pm$ 0.0142 \\
        \hline
        RI & No & \textbf{Yes} \\
        \hline
    \end{tabular}
    \caption{University\_v1 dataset aggregated table metrics.}
    \label{tab:Results-University_v1}
    
    \begin{tabular}{c | c | c}
        Metrics & HMA1 & HCTGAN \\
        \hline
        CS & \textbf{0.885} $\boldsymbol{\pm}$ \textbf{0.0009} & 0.785 $\pm$ 0.0148 \\
        CPT & \textbf{0.840} $\boldsymbol{\pm}$ \textbf{0.0012} & 0.716 $\pm$ 0.0118 \\
        PCR & \textbf{0.636} $\boldsymbol{\pm}$ \textbf{0.0014} & 0.619 $\pm$ 0.0002 \\
        RC & 0.951 $\pm$ 0.0015 & \textbf{0.997} $\boldsymbol{\pm}$ \textbf{0.0046} \\
        NRS & 0.554 $\pm$ 0.0014 &\textbf{0.609} $\boldsymbol{\pm}$ \textbf{0.0036} \\
        BA & \textbf{1.0} $\boldsymbol{\pm}$ \textbf{0.0} & 0.886 $\pm$ 0.0060 \\
        \hline
        RI & No & \textbf{Yes} \\
        \hline
    \end{tabular}
    \caption{Hepatitis\_std\_v1 dataset aggregated table metrics.}
    \label{tab:Results-Hepatitis_std_v1}
    
    \begin{tabular}{c | c | c}
        Metrics & HMA1 & HCTGAN \\
        \hline
        CS & 0.551 $\pm$ 0.0 & \textbf{0.593} $\boldsymbol{\pm}$ \textbf{0.0115} \\
        CPT & \textbf{0.576} $\boldsymbol{\pm}$ \textbf{0.0} & 0.543 $\pm$ 0.0069 \\
        PCR & \textbf{1.0} $\boldsymbol{\pm}$ \textbf{0.0} & \textbf{1.0} $\boldsymbol{\pm}$ \textbf{0.0} \\
        RC & 0.796 $\pm$ 0.0 & \textbf{0.980} $\boldsymbol{\pm}$ \textbf{0.0025} \\
        NRS & \textbf{1.0} $\boldsymbol{\pm}$ \textbf{0.0} & \textbf{1.0} $\boldsymbol{\pm}$ \textbf{0.0} \\
        BA & \textbf{1.0} $\boldsymbol{\pm}$ \textbf{0.0} & 0.702 $\pm$ 0.0155 \\
        \hline
        RI & \textbf{Yes} & \textbf{Yes} \\
        \hline
    \end{tabular}
    \caption{Pyrimidine\_v1s dataset aggregated table metrics.}
    \label{tab:Results-Pyrimidine_v1}
\end{table}

\subsection{Discussion}
\label{sec:Results_Discussion}

A clear trend is that the HMA1 model achieves slightly better quality scores (CS \& CPT), or performs almost equally to, the HCTGAN model on all three datasets \ref{tab:Results-University_v1}. However, sampling using the HMA1 model takes significantly longer time than with the HCTGAN algorithm, primarily due to the fact that HMA1 uses recursion to model relating tables. Furthermore, the HMA1 model does not guarantee referential integrity (RI) on the generated synthetic data. Although we can see that the HMA1 model manages to achieve referential integrity for the Pyrimidine\_v1 dataset, generating synthetic data that does not have referential integrity could potentially break important database systems if used without care. This is why we believe referential integrity is a critical attribute of synthetic data that should always be considered during evaluation, and is an aspect that has unfortunately so far been overlooked in prior research.

The HCTGAN sampling algorithm produced data that always covered the numerical and categorical ranges better than the HMA1 model. However, HCTGAN also produced data that not always adhered to the boundaries of the real data. One could attribute the good RC scores to the HCTGAN models being poor at following min- and max boundaries, which the results suggest. However, a dataset is usually only a subset of all possible and valid observations, so it does not mean that the data in reality would be bounded to the same min- and max values that the dataset is bounded by. Thus, by using the HCTGAN models for synthetic data generation one could find outliers and new types of samples that otherwise would be missed with the HMA1 model.

Both models tended to produce a similar amount of novel synthetic rows (NSR), but the HCTGAN algorithm consistently produced a larger percentage of novel rows. This could also be attributed to the fact that the HCTGAN algorithm produces data which can lie outside of the min- and max values present in the training dataset. The HMA1 model outperforms HCTGAN when it comes to modeling accurate parent child relationships (PCR). This was expected since the sampling algorithm in HCTGAN utilizes heuristics for estimating how many related child rows should be generated based on the number of parent rows.

\section{Conclusions} 
\label{sec:Conclusions}

In this paper we proposed the HCTGAN model for multi-tabular synthetic data generation. We conclude that the proposed model sufficiently well captures the characteristics of the real data, whilst the sampling algorithm also guarantees referential integrity in the generated synthetic data. The proposed sampling algorithm for HCTGAN allows it to efficiently sample large amounts of synthetic data for arbitrarily complex relational datasets, whereas the HMA1 algorithm is limited in this respect due to its recursive nature. Because the results indicate that the HMA1 model produces higher quality synthetic data, but does not scale for larger relational datasets, we suggest that the HCTGAN algorithm should be used whenever large amounts of synthetic data is needed, or whenever the dataset is deep and contains complex table relations.

For future work we propose to investigate the effect of the chosen parent column distribution which the child tables condition their generator on. One could estimate these distributions similarly to how the columns in the dataset are modelled using mode-specific normalization. This could potentially increase the quality of the generated synthetic data due to better capturing relating information between the tables. Furthermore, performing an ablation study on the conditioning of parent information would be insightful to determine its effect.

Finally, we conclude that more research in multi-tabular synthetic data algorithms is needed to establish a foundation for robust, and privacy preserving, synthetic data algorithms, where specific importance should be put on researching any privacy related issues with GAN based synthetic data models.

\bibliographystyle{plain}
\bibliography{kallor}

\newpage

\appendix

\section{HCTGAN training algorithm}
\label{appendix:HCTGAN-training-algorithm}

\begin{algorithm}[h]
\caption{HCTGAN training, one iteration}\label{algorithm:HCTGAN-train}
\begin{algorithmic}[1]
    \Require $\mathcal{G}$, $\mathcal{C}$, $\mathcal{S}$ \Comment{Generators, critics, and data samplers}
    \Ensure $\exists\ \mathcal{T}_i\ :\ |\mathcal{P}(\mathcal{T}_i)| \geq 1 $ \Comment{At least one table relation has to exist}

    \Procedure{train}{$\mathcal{G},\ \mathcal{C},\ \mathcal{S}$}
    \For{$\mathcal{T}_i$ in sorted tables}
        \For{discriminator step}
            \If{$|\mathcal{P}(\mathcal{T}_i)| \geq 1$}
                \State $\mathbf{z} \gets \mathbf{z} \sim \biggl( \mathcal{N}(\boldsymbol{\mu}_{p,1}, \boldsymbol{\sigma}_{p,1}),\ \dots,\ \mathcal{N}(\boldsymbol{\mu}_{p,n}, \boldsymbol{\sigma}_{p,n}) \biggr)$ \Comment{Sample noise from parent Gaussians}
            \Else
                \State $\mathbf{z} \gets \mathbf{z} \sim \mathcal{N}(0, 1)$
            \EndIf
            \State $\mathbf{c}, \mathbf{m} \gets \mathcal{S}_i(\mathcal{T}_i)$ \Comment{Sample conditional- and mask vector}
            \State $\mathbf{z} \gets \mathbf{z} \oplus \mathbf{c}$
            \State $\mathbf{\Tilde{x}} \gets \mathcal{G}_i(\mathbf{z})$ \Comment{Generate synthetic data from $\mathbb{P}_{\mathcal{G}}$}
            \State $\mathbf{x} \gets \mathcal{S}_i(\mathcal{T}_i)$ \Comment{Sample real data from $\mathbb{P}_x$}
            \State $\mathcal{L}_{\mathcal{C}} \gets \mathbb{E}[\mathcal{C}_i(\mathbf{\Tilde{x}})] - \mathbb{E}[\mathcal{C}_i(\mathbf{x})]$ \Comment{Wasserstein loss for critic}
            \State $\mathcal{L}_{GP} \gets \mathbb{E}[(\| \nabla_{\mathbf{\hat{x}}}\ \mathcal{C}(\mathbf{\hat{x}}) \|_2 - 1)^2]$ \Comment{Gradient penalty with packing}
            \State $\mathcal{C}_i \gets \mathcal{C}_i - \eta\cdot\text{Adam}(\mathcal{L}_{\mathcal{C}} + \lambda\mathcal{L}_{GP})$ \Comment{Optimize critic with added gradient penalty}
        \EndFor
        \State Sample $\mathbf{z}$, $\mathbf{\hat{x}}$ and $\mathbf{m}$ following lines $4$ to $11$
        \State $\mathcal{L}_{\mathcal{G}} \gets -\mathbb{E}[\mathcal{C}_i(\mathbf{\Tilde{x}})]$
        \State $\mathcal{G}_i \gets \mathcal{G}_i - \eta\cdot\text{Adam}(\mathcal{L}_{\mathcal{G}} + \text{cross entropy}(\mathbf{c}, \mathbf{m}))$ \Comment{Optimize generator with added cross entropy}
        \State $\boldsymbol{\mu}_i \gets \biggl( \frac{1}{N}\sum_i^N\mathbf{\hat{x}}_1,\ \dots,\ \frac{1}{N}\sum_i^N\mathbf{\hat{x}}_d  \biggr)$ \Comment{Column-wise mean}
        \State $\boldsymbol{\sigma}_i \gets \biggl( \sqrt{\frac{1}{N - 1}\sum_i^N(\mathbf{\hat{x}}_1 - \Bar{\mathbf{x}}_1)},\ \dots,\ \sqrt{\frac{1}{N - 1}\sum_i^N(\mathbf{\hat{x}}_d - \Bar{\mathbf{x}}_d)} \biggr)$ \Comment{Column-wise standard deviation}
    \EndFor
    \State \textbf{return} $\mathcal{G}$
    \EndProcedure

\end{algorithmic}
\end{algorithm}

\newpage

\section{HCTGAN sampling algorithm}
\label{appendix:HCTGAN-sampling-algorithm}

\begin{algorithm}[h]
\caption{HCTGAN sampling, relational}\label{algorithm:sample}
\begin{algorithmic}[1]
    \Require $\mathcal{G}$,\ $\Gamma$,\ $\mathcal{S},\ n$ \Comment{Trained generators, id distributions, data samplers, and number of samples}
    \Ensure $\exists\ \mathcal{T}_i\ :\ |\mathcal{P}(\mathcal{T}_i)| \geq 1 $

    \Procedure{sample}{$\mathcal{G},\ \Gamma,\ \mathcal{S},\ n$}
    \State synthetic\_data $\gets \emptyset$
    \State ids $\gets \emptyset$
    \For{$\mathcal{T}_i$ in sorted tables}
        \State num\_foreign $\gets \emptyset$
        \For{$\mathcal{T}_p$ in $\mathcal{P}(\mathcal{T}_i)$}
            \If{relationship type$(\mathcal{T}_p) == $ one-to-one}
                \State num\_foreign[$\mathcal{T}_p$] $\gets |\text{ids[}\mathcal{T}_p\text{]}|$ \Comment{Append parent amount to current table}
            \Else
                \State num\_foreign[$\mathcal{T}_p$] $\gets \Gamma(\mathcal{T}_p, \mathcal{T}_i)$ \Comment{Sample amount from learned Gamma distribution}
            \EndIf
        \EndFor
        \State num\_primary $\gets n$
        \State sampled\_ids $\gets \emptyset$
        \If{$|\mathcal{P}(\mathcal{T}_i)| \geq 1$}
            \If{one-to-one $\in$ relationship types$(\mathcal{P}(\mathcal{T}_i))$} \Comment{Special case for improved cardinality shape}
                \For{$\mathcal{T}_p$ in $\mathcal{P}(\mathcal{T}_i)$}
                    \State sampled\_ids[$\mathcal{T}_p$] $\gets$ ids[$\mathcal{T}_p$] \Comment{Append all foreign keys to current table}
                    \State num\_primary $\gets |$sampled\_ids[$\mathcal{T}_p$]$|$ \Comment{Take either parents amount, they are the same}
                \EndFor
            \Else
                \State num\_keys $\gets \text{max}(\text{num\_foreign})$
                \For{$\mathcal{T}_p$ in $\mathcal{P}(\mathcal{T}_i)$}
                    \State parent\_ids $\gets$ ids[$\mathcal{T}_p$]
                    \State sampled\_foreign $\gets$ randomly pick num\_keys from ids[$\mathcal{T}_p$] \Comment{With replacement}
                    \While{parent\_ids $\setminus$ sampled\_foreign $\neq \emptyset$} \Comment{Ensures referential integrity}
                        \State sampled\_foreign $\gets$ randomly pick num\_keys[$\mathcal{T}_p$] from ids[$\mathcal{T}_p$]
                    \EndWhile
                    \State sampled\_ids[$\mathcal{T}_p$] $\gets$ sampled\_foreign
                \EndFor
                \State num\_primary $\gets$ num\_keys
            \EndIf
        \EndIf
        \State ids[$\mathcal{T}_i$] $\gets \{0,\ \dots,\ \text{num\_primary}\}$ \Comment{Create unique primary keys}

        \State Generate $\mathbf{\Tilde{x}}$ following lines $4$ to $11$ in \ref{algorithm:HCTGAN-train} \Comment{Generate synthetic row}
        \State $\mathbf{\Tilde{x}} \gets \mathbf{\Tilde{x}}$ + ids[$\mathcal{T}_i$] \Comment{Append primary keys to current table}
        \For{$\mathcal{T}_p$ in $\mathcal{P}(\mathcal{T}_i)$}
            \State $\mathbf{\Tilde{x}} \gets \mathbf{\Tilde{x}} +$ ids[$\mathcal{T}_p$] \Comment{Append foreign keys to current table}
        \EndFor
        \State synthetic\_data[$\mathcal{T}_i$] $\gets \mathbf{\Tilde{x}}$
    \EndFor
    \State \textbf{return} synthetic\_data
    \EndProcedure
\end{algorithmic}
\end{algorithm}

\end{document}